\documentclass{article}

\usepackage[preprint]{neurips_2021}

\usepackage{graphicx}
\usepackage{natbib}
\usepackage{enumitem}
\usepackage{wrapfig, sidecap}
\usepackage[utf8]{inputenc} 
\usepackage[T1]{fontenc}    
\usepackage{hyperref}       
\usepackage{url}            
\usepackage{booktabs}       
\usepackage{amsfonts}       
\usepackage{nicefrac}       
\usepackage{microtype}      
\usepackage{xcolor}         
\usepackage[ruled,vlined]{algorithm2e}

\title{Selecting for Selection:\linebreak Learning To Balance Adaptive and Diversifying Pressures in Evolutionary Search}

\author{Kevin Frans$^{1,2}$,  L.\ B.\ Soros$^2$ and  Olaf Witkowski$^{2,3,4}$ \\
\mbox{}\\
$^1$Massachusetts Institute of Technology, Cambridge, MA, USA \\
$^2$Cross Labs, Cross Compass Ltd., Tokyo, Japan\\
$^3$Earth-Life Science Institute, Tokyo Institute of Technology, Japan\\
$^4$College of Arts and Sciences, University of Tokyo, Japan\\

kvfrans@csail.mit.edu}

\begin{document}

\maketitle

\begin{abstract}
Inspired by natural evolution, evolutionary search algorithms have proven remarkably capable due to their dual abilities to radiantly explore through diverse populations and to converge to adaptive pressures. A large part of this behavior comes from the \emph{selection function} of an evolutionary algorithm, which is a metric for deciding which individuals survive to the next generation. In deceptive or hard-to-search fitness landscapes, greedy selection often fails, thus it is critical that selection functions strike the correct balance between gradient-exploiting adaptation and exploratory diversification. This paper introduces Sel4Sel, or Selecting for Selection, an algorithm that searches for high-performing neural-network-based selection functions through a meta-evolutionary loop. Results on three distinct bitstring domains indicate that Sel4Sel networks consistently match or exceed the performance of both fitness-based selection and benchmarks explicitly designed to encourage diversity. Analysis of the strongest Sel4Sel networks reveals a general tendency to favor highly novel individuals early on, with a gradual shift towards fitness-based selection as deceptive local optima are bypassed.
\end{abstract}

\section{Introduction}



    



Evolution in nature has inspired a wide variety of applications in computational domains, from open-ended generative systems to objective-based optimization algorithms. A key factor in the strength of evolutionary algorithms are their dual abilities to radiantly explore and to converge to adaptive pressures, both of which are important aspects of \emph{evolvability} -- often described as a population’s “capacity to evolve” \citep{veenstra:thesis}, but generally “the ability of a biological system to produce phenotypic variation that is both heritable and adaptive” \citep{dawkins:alife88,crother:ee19,payne:nature19}. In fact, the evolution of evolvability has been recognized as a kind of open-ended evolution \citep{packard:alife19}, which itself may be a key to open-ended intelligence \citep{stanley:oreilly}.

A key observation is that the evolvability of an evolutionary algorithm can be quantified, and thus optimized, through a meta-learning loop. Specifically, this work quantifies the evolvability of an evolutionary algorithm as the expected fitness of its resulting population after finite evolution in a novel environment. By running a meta-evolutionary loop that evolves the \emph{evolutionary algorithm} itself, aspects of evolution can be optimized, revealing insight on which evolutionary behaviors consistently improve performance and learning ability.

Solving optimization problems in particular can be challenging if fitness landscapes are sufficiently deceptive, i.e. containing many local optima and thereby requiring traversing valleys to reach the peaks. Simple greedy (i.e. purely adaptive) algorithms will fail to maximize them. Thus, it is critical that evolutionary algorithms applied to these problems strike the correct balance between selecting for gradient-exploiting adaptation versus exploratory diversification in order to bypass local optima.

This work focuses specifically on meta-evolution of selection functions in tournament-based genetic algorithms applied to optimization problems. Generally, a genetic algorithm maintains a population of individuals, which are iteratively mutated and then selectively culled according to some selection function. By adjusting this selection function, the overall behavior of a genetic algorithm can be directly manipulated.

Presented in this paper is a novel algorithm called \emph{Sel4Sel}, or \emph{Selecting for Selection}, which aims to discover neural-network-based selection functions that rank individuals based on characteristics such as fitness, novelty, and age. Once optimized through a meta-evolutionary loop, Sel4Sel networks learn to select individuals so as to optimally balance pressures for exploration and exploitation during evolutionary search. The selection behaviors discovered through Sel4Sel match or outperform the strongest baselines on three bitstring domains with qualitatively different fitness landscapes, showing that meta-evolution can consistently discover high-performing selective behavior. The behaviors of the strongest Sel4Sel networks are then examined, revealing that optimal selection involves rewarding highly novel individuals early on, then swapping to a greedy fitness-based selection in later stages.

The main contributions of this work are:
\begin{itemize}
    \item Quantifying the evolvability of evolutionary algorithms as a measurable meta-learning objective; namely, the expected fitness of the population after evolving for a number of generations in a novel environment.
    
    \item Presenting an algorithm, \emph{Selecting for Selection (Sel4Sel)}, which discovers neural-network-based selection functions resulting in high-performing evolutionary algorithms.
    
    \item Examining the behavior of strong Sel4Sel models to illuminate the utility of various input metrics in ranking individuals, as well as revealing exploration/exploitation tradeoff patterns.
\end{itemize}

\begin{figure}[t!]
\center{
    \includegraphics[width=0.9\linewidth]{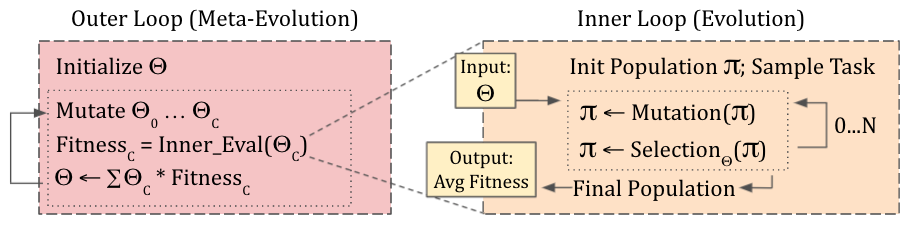}
    }
\caption{\textbf{Outline of Sel4Sel.} The Sel4Sel algorithm runs a meta-evolutionary loop that evolves the \emph{selective behavior} of an inner evolutionary algorithm, represented by neural network $\theta$. For each meta-evolutionary iteration, various selection function mutations $\theta_0$...$\theta_c$ are evaluated by fully running an inner evolutionary algorithm, starting from random initial population $\pi$ and evolving for $N$ generations. Once trained, Sel4Sel discovers selective behaviors which dynamically balance between exploration and exploitation, enabling efficient adaptation on the tasks at hand.
}
\label{fig:network}
\end{figure}

\section{Related work}

\textbf{Adaptation and Diversification in Evolutionary Search.} Evolutionary computation has historically focused on objective-based optimization algorithms, where selection was based primarily on minimizing distance between members of a population and some optimum in the search space \citep{dejong:book02}. The advent of novelty search \citep{lehman:ec11}, which instead selects for individuals that maximize diversity in the population irrespective of the global optimum, represented a radical departure from the traditional approach. Since then, a variety of algorithms have been developed to combine selection for both fitness and diversity, including novelty search with local competition \citep{lehman:gecco2011} and MAP-Elites \citep{cully:nature15}. The latter algorithms largely form the basis of the burgeoning class of quality diversity (QD) algorithms, which discover a diverse set of high-performing individuals in a given search space \citep{pugh:frontiers16}. However, most QD algorithms require manual specification of a phenotypic feature vector that encodes the dimensions along which diversity of high-performing individuals is explicitly maintained. Recent work on QD algorithms has resulted in methods that automate configuration for that particular subclass of evolutionary algorithms. In particular, \cite{cully:gecco19} and \cite{bossens:gecco20} use PCA and CMA-ES, respectively, to construct feature descriptors without human intervention. 

This work views diversity, along with other properties, not as human-induced measurements but as a byproduct of meta-evolution. For what purpose is diversity introduced to evolutionary search? A reasonable intuition is that diverse populations have a better chance to adapt to novel challenges. In Sel4Sel, the meta-learning objective of adapting to novel challenges is directly optimized, allowing evolution to discover useful behaviors on its own. Thus, diversity is not an externally imposed requirement, but something that should be discovered through meta-evolution if its presence is indeed useful. On a broad level, the aim of this work is to bridge the many studies of what evolutionary algorithms should optimize for, e.g. fitness, novelty, etc. by directly measuring their effects on long-term learning.

\textbf{Meta Learning.} Because evolutionary algorithms can be seen as a kind of learning system, this work on learning evolutionary algorithms also relates to the field of meta-learning, or learning to learn \citep{learn2learn}. In traditional meta-learning, an outer loop optimizes the behavior of an inner loop, which often aims to quickly learn to solve new tasks \citep{metasurvey}. Evolutionary methods have traditionally been used in the outer loop as black box optimizers for the parameters of an inner loop learning algorithm, such as neural network structures \citep{evolvestruct} or policy gradient algorithms \citep{epg}. Work that focuses on improving evolutionary search generally propose online statistical methods, such as CMA-ES \citep{cmaes} or variants of self-adaptation \citep{selfadaptation}, although meta-evolutionary loops have been used to discover rich behavioral spaces \citep{bossens:gecco20}. In Sel4Sel, both inner and outer loops are evolutionary algorithms, with the goal to optimize the selective behavior of the inner evolutionary algorithm towards achieving high fitness in the long run.





\section{Sel4Sel Algorithm}

Selecting For Selection (Sel4Sel; Algorithm \ref{alg:sel4sel}) aims to balance evolutionary search objectives to generate high-performing individuals in tournament-based genetic algorithms. In a genetic algorithm, a population of individuals are iteratively mutated, then a subset of the population is selected for reproduction according to an objective function (usually simply the domain's objective function). By adjusting this selection function, the behavior of the genetic algorithm can be consistently controlled.

\begin{algorithm}[t!]
\SetAlgoLined
\DontPrintSemicolon
\textbf{Input:} Initial network parameters $\theta$, Task distribution $T$.\;
\textbf{Outer Loop:} \;
\For{iteration=0...I} {
    \For{copy=0...C} {
        \textbf{Inner Loop:} \;
        $\theta_c$ = $\theta$ + Normal() * $\sigma$\;
        Sample task $t \sim T$; Initialize GA with population $\pi^c$.\; 
        \For{generation=0...$N$}{
            Create offspring population. $\phi^c_p \leftarrow$ Mutate($\pi^c_p$) \;
            Compare each offspring $\phi^c_p$ against a random competitor from the population $\pi^c_r$. \;
            \If{$ \textnormal{InternalFitness}(\theta_c, \phi^c_p) >= \textnormal{InternalFitness}(\theta_c, \pi^c_r) $} {
                Replace competitor with offspring. $\pi^c_r \leftarrow \phi^c_p$ \;
            }
        }
    }
        Update $\theta$ towards rank-weighted average of copies. $\theta \leftarrow \sum_{c=0}^{C} \theta_c *$ Rank(Mean(Fitness($\pi^c$)))
}
\caption{Selecting for Selection (Sel4Sel)}
\label{alg:sel4sel}
\end{algorithm}

In Sel4Sel, a neural network encodes the selection function, represented by the parameters $\theta$. Inputs include various measurements about an individual, such as underlying/objective fitness, age, or novelty, and output is a scalar internal fitness. This internal fitness is then used during tournament selection, whereby offspring replace members of the population with lower internal fitness scores. 

On a high level, the goal of Sel4Sel is to optimize the selection function $\theta$ such that the internal genetic algorithm performs strongly. This goal is accomplished through an inner and outer loop. The inner loop serves as an evaluation function, which measures the strength of a genetic algorithm run using selection function $\theta$. The outer loop then iteratively searches for $\theta$ parameters which display high performance, repeated until convergence.

In the inner loop, Sel4Sel measures the performance of a given selection function by fully running a genetic algorithm and evaluating the final population. First, a task is sampled from task distribution $T$, and a population of $P$ individuals are randomly initialized. A tournament-based genetic algorithm is then run for $N$ generations on the sampled task. Every generation, each individual produces a slightly mutated offspring, which is then compared against a random competitor from the population. Of these two individuals, the one with higher internal fitness -- as defined by selection function $\theta$ -- remains within the population. After $N$ generations, the overall performance of selection function $\theta$ is measured by taking the average underlying fitness of the final population.

In the outer loop, Sel4Sel searches for $\theta$ parameters that result in high-performing selection functions. While any black box optimization method can suffice, this work makes use of the evolutionary strategies method detailed in \cite{salimans:es}. Specifically, $C$ copies of the base neural network are created, each with a random amount of noise added to its parameters. The performance of each copy is measured by fully running an inner loop genetic algorithm, after which the base network is updated towards a performance-weighted average of the copies.

\subsection{Network Structure and Input Metrics}

The neural network used to encode selection functions in Sel4Sel computes an internal fitness score for every individual in a population based on a variety of evolutionary features. As shown in Figure \ref{fig:network}, the network encoding is comprised of three feedforward layers in sequence. Inputs to the neural network are a set of evolutionary features collected about an individual. Many of these metrics considered have been used independently or in some limited combination as objectives in evolutionary search, as described in Related Work. Underlying Fitness directly measures the fitness returned by the task domain, whereas Rank returns the normalized rank of an individual compared to its population. Age represents the number of generations the individual has remained in the population. Novelty measures the distance of an individual’s genome from its nearest neighbors, as adopted from Novelty Search \citep{lehman:ec11}. In this work, novelty is measured through the sum of Hamming distances between an individual’s five nearest neighbors. Finally, Noise is an input which is set to a random number between 0 and 1, and Generation returns a normalized count of how many generations the genetic algorithm has been running for. For more implementation details, refer to the source code of the experiments linked below\footnote{Source code for domains and algorithms is available at: https://github.com/kvfrans/sel4sel
}.


\begin{figure}[t!]
\center{
    \includegraphics[width=0.9\linewidth]{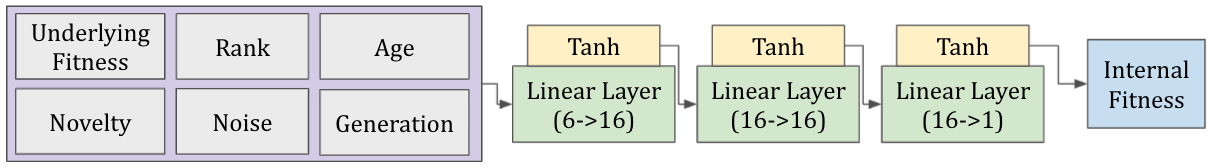}
    }
\caption{\textbf{Neural Network structure encoding selection functions for Sel4Sel.} Evolutionary features are collected for each individual, which are passed as inputs to the selection function network to produce the individual's internal fitness score. The parameters for this network are meta-learned by evolution strategies to produce a high-performing evolutionary algorithm. Section 3.1 describes the inputs to the network in detail.}
\label{fig:network}
\end{figure}

\section{Methodology}

The aim of this work is to examine what kinds of evolutionary search objectives (pressures for selection based on e.g.\ fitness and diversity), when used in tournament-based genetic algorithms, consistently generate high-performing populations. The main avenue for exploring this question is Sel4Sel, an algorithm which searches for high-performance selection functions through a meta-evolutionary loop. In a series of experiments, Sel4Sel is shown to consistently outperform human-designed selection functions from literature. A later section examines the strongest Sel4Sel networks to shed light on what selective behaviors are generally useful.

Experiments compare novel selection functions learned through Sel4Sel against baseline selection functions from literature that have been explicitly designed to encourage both fitness-based adaptation and diversification. Importantly, evolvability requires both of these pressures, yet surprisingly most quantitative studies of evolvability in artificial life focus on either evolvability as adaptation \citep{medvet:gecco17, veenstra:alife20} or evolvability as diversification \citep{mengistu:gecco16, gajewski:gecco19}, but not both. Baseline comparisons are chosen to represent various methods of encouraging evolvability, without explicitly requiring both adaptation and diversification, so as to remain agnostic to the ideal balance between the two.

Comparisons are presented in three bitstring domains with qualitatively different fitness landscapes, each providing distinct challenges. As demonstrated in experiments below, final populations produced via Sel4Sel match or outperform the strongest baselines in every domain.

In all experiments, the full Sel4Sel algorithm is first run until completion on each domain. Specifically, the selection function $\theta$ meta-evolves in the outer loop for 10,000 iterations, with 20 copies evaluated each time. Each inner loop evaluation samples a random starting population, along with a random task from the domain. Note that meta-evolutionary loop efficiency is not a focus herein, and in general the network converges much earlier than the full iteration count. All experiments run on within a few hours on a 4-CPU laptop.

Once the Sel4Sel meta-evolution has finished, the resulting $\theta_\emph{final}$ parameters represent the selection function to be evaluated in test-time. For test-time comparisons, which are presented below, a random starting population and random task are sampled from the domain. A full genetic algorithm is then run using each selection function; namely the learned Sel4Sel metrics, along with baseline metrics from literature. This is repeated 20 times for each trial, with the results averaged. In all trials, the population size of the genetic algorithm is 50, and evolution runs for 2000 generations, following parameter choices from \cite{veenstra:alife20}.

\subsection{Domains}
The domains presented represent three task distributions with qualitatively different fitness landscapes, to test the robustness of various selection functions. In each domain, an individual is represented by a 16-bit binary string $B$. During mutation, offspring are produced by copying a parent individual’s bitstring, along with a 0.05 chance for each bit to turn into a random bit. In all domains, the fitness function has a minimum of 0 and a maximum of 32. Note that the learning problems presented in each domain are best viewed as “transform a population of random individuals into a population of high-fitness individuals”. Thus, while the fitness function is static in the first and third domains, a distribution of tasks is presented as the starting population is randomized, and mutation behavior carries intrinsic noise. The following domains are considered:

\textbf{Convex Bits.} The first domain, Convex Bits, represents a straightforward task in which greedy search performs well. The fitness of an individual is simply the sum of its bits multiplied by two. A strong selection function should iteratively select for individuals with higher underlying fitness.

\noindent
\begin{equation}
\textnormal{Fitness}(B) = \sum\nolimits_{b=0}^{15} B_b * 2
\end{equation}

\textbf{Hashed Bits.} The second domain, Hashed Bits, represents a task that requires explicit exploration. The fitness of an individual is $2^R$, where $R$ is a random number between 0 and 5 generated with the constant-shifted bitstring as a seed. Note that this constant is only changed when the genetic algorithm ends; i.e.\ the fitness of an individual will remain the same for the duration of the current evolutionary run, but all fitnesses will be randomized in the next run. Notably, in this domain there is no correlation between the fitness of an individual and its offspring, or any related bitstring. Thus, a strong selection function should learn a method to intrinsically explore the space of bitstrings, while exploiting high-fitness individuals.

\noindent
\begin{equation}
\textnormal{Fitness}(B) = 2\textsuperscript{Min(Abs(Normal(0, 4/3), 5)}
\end{equation}

\begin{SCfigure}
    \includegraphics[width=0.5\linewidth]{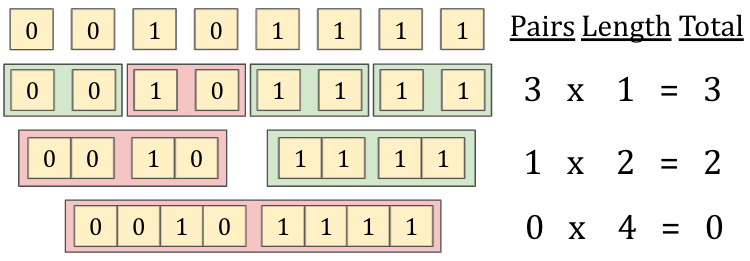}
\caption{\textbf{Fitness calculation for the Deceptive Bits domain.} Fitness is measured by recursively computing the number of pairs at every length, with each pair awarding fitness equal to its length. In this example, there are 3 pairs of length 1, 1 pair of length 2, and 0 pairs of length 4; awarding a total fitness of $3*1 + 1*2 + 0*4 = 5$.}
\label{fig:hierarchicaliff}
\end{SCfigure}

\textbf{Deceptive Bits.} The third domain, Deceptive Bits, is a task wherein the fitness function is inherently deceptive. Fitness is calculated recursively by comparing pairs of bits, then pairs of pairs of bits, etc., until the entire left side of the bitstring is being compared to the entire right side. This is otherwise known as the hierarchical if-and-only-if function as detailed by \cite{veenstra:alife20}. An illustration is shown in Figure \ref{fig:hierarchicaliff}. In this domain, there are numerous local maxima in which a greedy algorithm may be trapped, thus a strong selection function should learn a way to escape such areas.

\subsection{Selection Functions}
In the presented experiments, learned Sel4Sel metrics are compared with various baseline metrics from literature. All comparisons are run via a tournament-based genetic algorithm, where individuals are selected according to their internal fitness, as defined by the chosen metric. Baseline metrics were chosen to create a spectrum of ability to encourage evolvability, namely:

\begin{itemize}
    \item \textbf{Underlying Fitness}, where internal fitness is the underlying task fitness function. Used as a selection function, this is equivalent to performing greedy selection towards high-fitness individuals and is the typically used in evolutionary algorithms with elitism \citep{dejong:book02}.
    
    \item \textbf{Novelty}, where internal fitness is the distance between an individual and its five closest neighbors. This work uses a Hamming distance, which measures the number of changes needed to transform one bitstring into the other. As a selection function, this corresponds to Novelty Search (NS) \citep{lehman:ec11}. Note that unlike canonical NS, novelty herein is calculated with respect to the current population without incorporating an archive.
    
    \item \textbf{Minimal Criterion}, where internal fitness is the underlying fitness, but capped at 16. This is equivalent to initially bootstrapping individuals to 16 fitness through greedy selection, followed by running Minimal Criterion selection \citep{soros:alife14, brant:gecco17}, where \emph{all} individuals reproduce as long as they exceed a fitness threshold.
    
    \item \textbf{Random Drift}, where internal fitness is a uniform random variable. This is equivalent to genetic drift algorithms, in which populations evolve without significant selective pressure. 
    
\end{itemize}







\section{Results}

\begin{wraptable}{r}{0.5\linewidth}
\center{
\begin{tabular}{|c||c|c|}\hline
Algorithm & Final Fitness & Final Novelty\\ \hline\hline
\textbf{Sel4Sel} & \textbf{32.0 $\pm$ 0.0} & 0.0 $\pm$ 0.0 \\
\textbf{Underlying Fit.} & \textbf{32.0 $\pm$ 0.0} & 0.0 $\pm$ 0.0 \\
Novelty & 15.84 $\pm$ 0.99 & 7.59 $\pm$ 0.49 \\
Min Crit. & 19.66 $\pm$ 1.02 & 3.94 $\pm$ 0.67 \\
Random Drift & 16.12 $\pm$ 1.87 & 3.07 $\pm$ 0.66 \\ \hline
\end{tabular}
}
\caption{\textbf{Convex Bits:} Average fitness and novelty of final populations, using various algorithms.} \label{table:convex_numbers}
\end{wraptable}

\begin{figure}[t!]
    \includegraphics[width=0.5\linewidth]{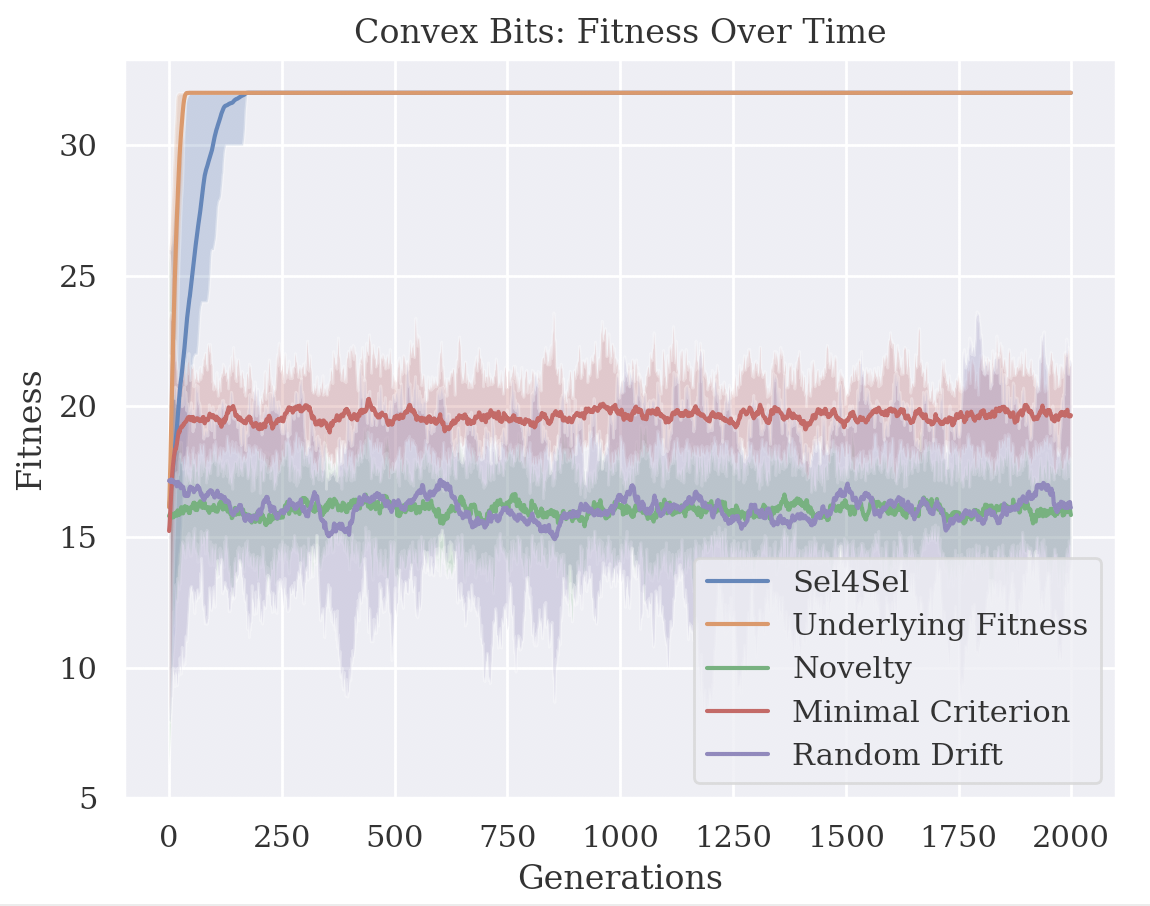}
    \includegraphics[width=0.5\linewidth]{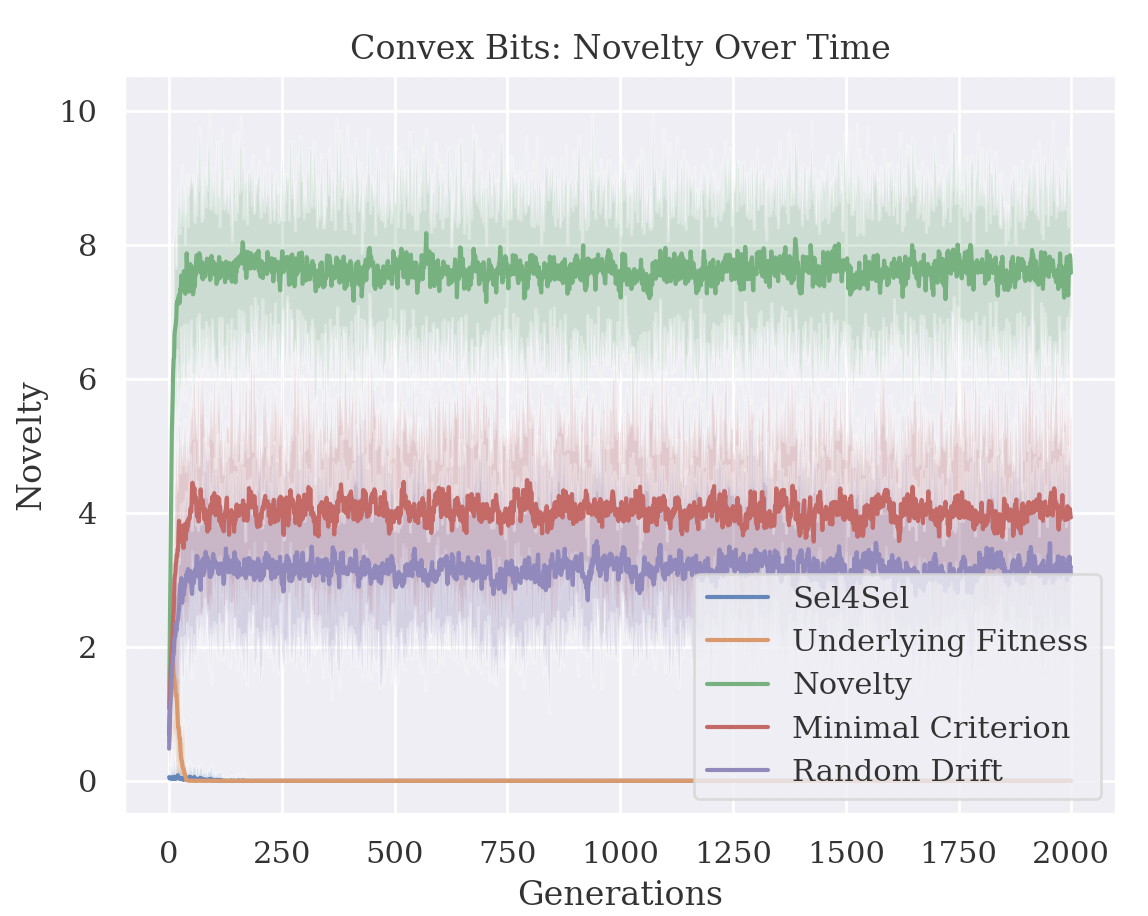}
\caption{The \textbf{Convex Bits} domain presents a simple fitness landscape where greedy selection works well. Sel4Sel learns to select for individuals with high underlying fitness, matching the strongest baseline.}
\label{fig:convex_fit}
\end{figure}

\textbf{Convex Bits.} This domain presents a smooth fitness landscape. Thus, greedy optimization with respect to underlying fitness yields strong results, as shown in Table \ref{table:convex_numbers}. The goal for Sel4Sel is to replicate this behavior. As displayed in Figure \ref{fig:convex_fit}, Sel4Sel successfully learns a selection function that rewards high-fitness individuals, consistently creating a final population with the maximum fitness of 32. In contrast, selection functions designed to nable diversity, such as minimal criterion and novelty, perform poorly. While novelty levels increase, indicating diverse populations, they do not lead to high fitness.

\begin{figure}[t!]
    \includegraphics[width=0.5\linewidth]{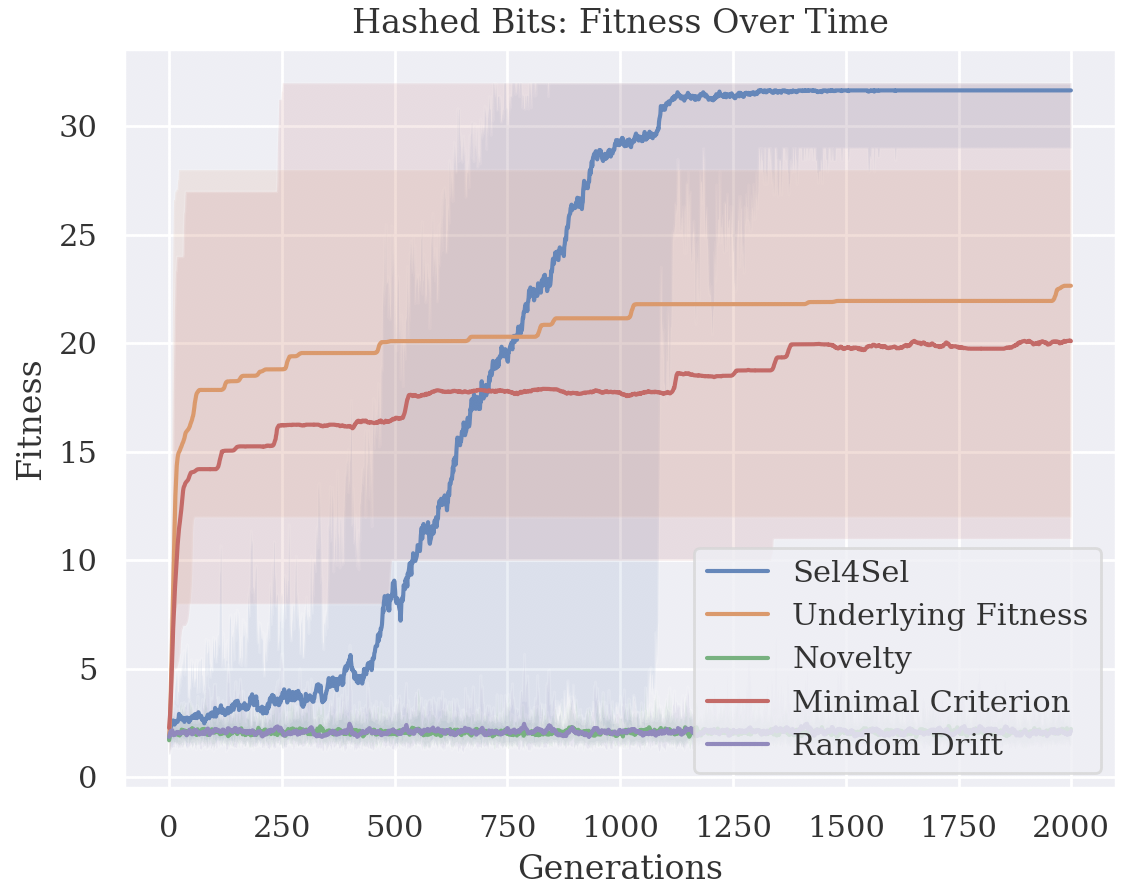}
    \includegraphics[width=0.5\linewidth]{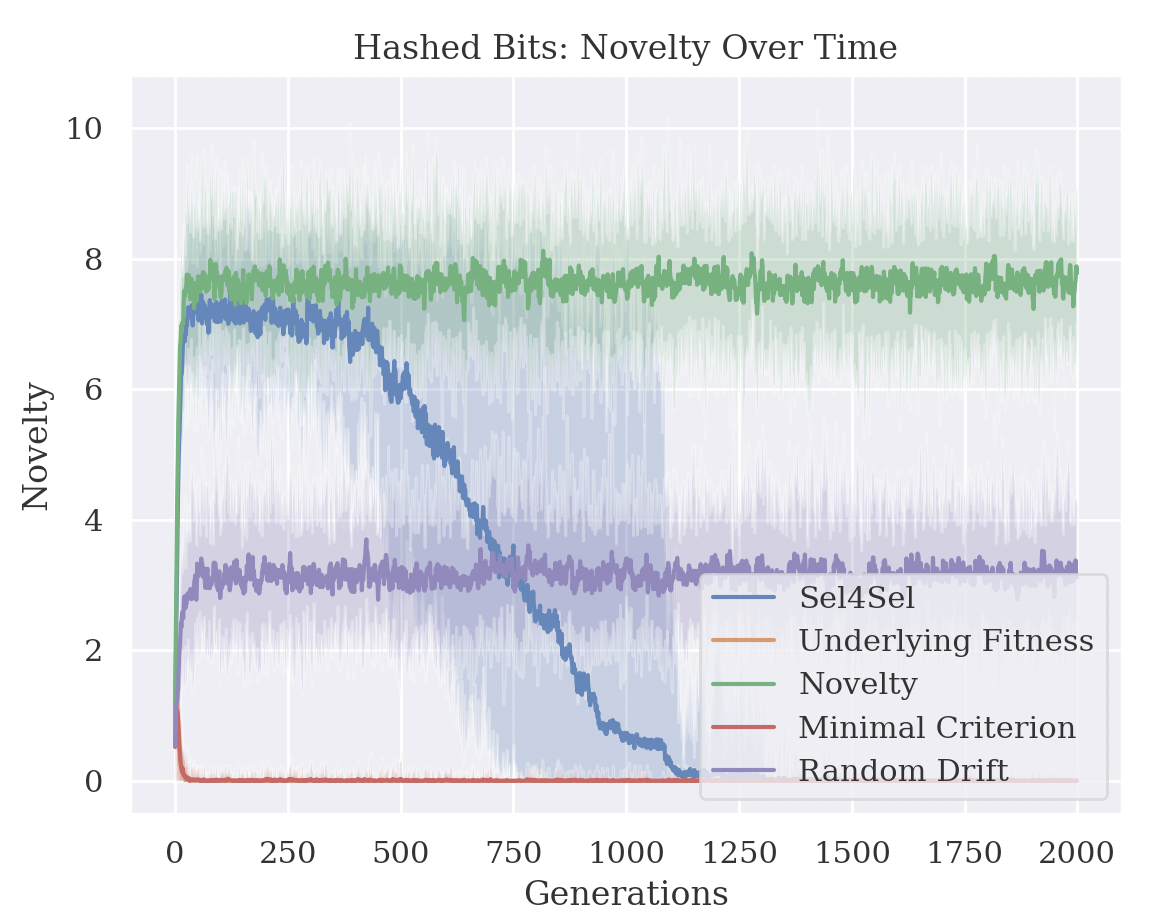}
\caption{The \textbf{Hashed Bits} domain presents a random fitness landscape which requires exploration. Sel4Sel learns to explore in early generations by encouraging novelty, only switching to exploitation when a maximum-fitness individual is located.}
\label{fig:hashed_fit}
\end{figure}

\begin{figure}[t!]
    \includegraphics[width=0.5\linewidth]{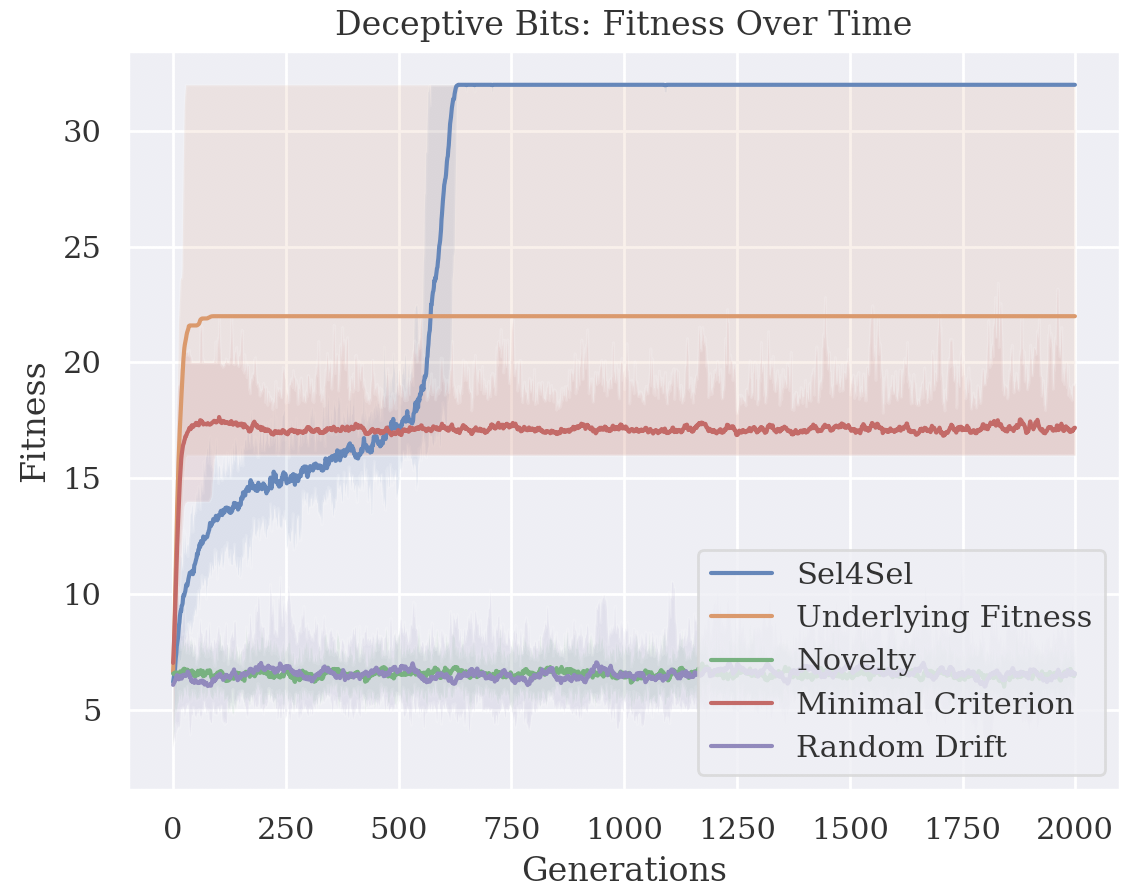}
    \includegraphics[width=0.5\linewidth]{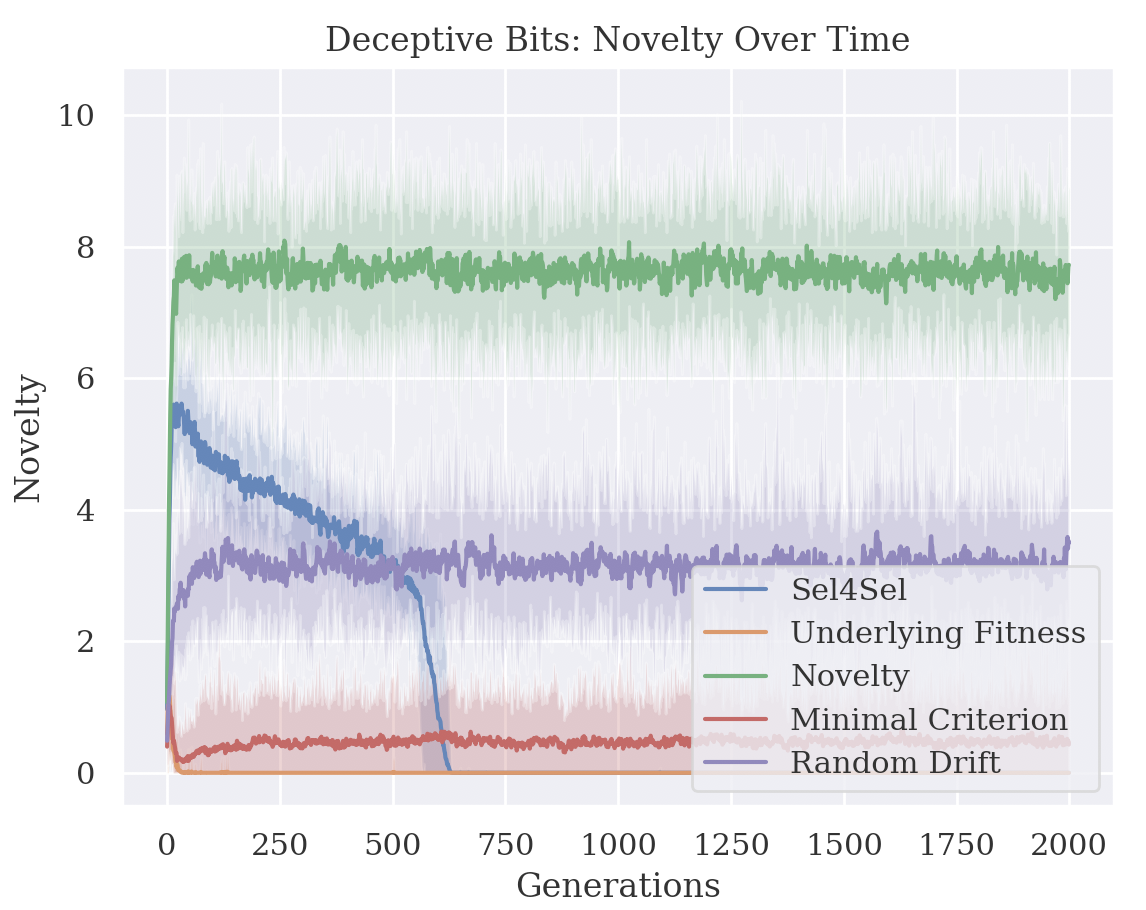}
\caption{The \textbf{Deceptive Bits} domain presents a deceptive fitness landscape with many local maxima. Sel4Sel learns to escape these maxima by selecting for a diverse population early on, which has a lower chance of becoming trapped when optimizing for underlying fitness.}
\label{fig:deceptive_fit}
\end{figure}


\begin{wraptable}{r}{0.5\linewidth}
\center{
\begin{tabular}{|c||c|c|}\hline
Algorithm & Final Fitness & Final Novelty\\ \hline\hline
\textbf{Sel4Sel} & \textbf{31.65 $\pm$ 0.73} & 0.0 $\pm$ 0.0 \\
Underlying Fit. & 22.65 $\pm$ 5.01 & 0.0 $\pm$ 0.0 \\
Novelty & 2.23 $\pm$ 0.35 & 7.78 $\pm$ 0.64 \\
Min Crit. & 20.1 $\pm$ 6.41 & 0.01 $\pm$ 0.03 \\
Random Drift & 2.13 $\pm$ 0.43 & 3.12 $\pm$ 0.53 \\ \hline
\end{tabular}
}
\caption{\textbf{Hashed Bits:} Average fitness and novelty of final populations, using various algorithms.} \label{table:hashed_numbers}
\end{wraptable}

\textbf{Hashed Bits.} In this domain, each bitstring is assigned a random fitness value. Due to the lack of correlation between neighboring bitstrings, direct optimization towards fitness does not lead to the global maximum. As shown in Table \ref{table:hashed_numbers}, the average result when greedily selecting for underlying fitness is considerably lower than the maximum. Even the minimum criterion objective, which is known to increase variation in populations, fails to reach the top.

A successful algorithm should instead intrinsically explore the bitstring space to find high-fitness individuals. Once a high-fitness individual is located, the selection function should then exploit this individual and allow it to spread through the population to maintain a high fitness average. This balance between exploration and exploitation (i.e.\ when to stop prioritizing diversity) should be learned through Sel4Sel meta-evolution.

Figure \ref{fig:hashed_fit} depicts Sel4Sel balancing exploration and exploitation. In early generations, Sel4Sel populations display high novelty, almost matching the levels of the pure novelty-optimizing population. This corresponds to a period of high exploratory pressure. Around generation 500, however, novelty levels drop, precisely when fitness levels begin to rise. This is when Sel4Sel begins to switch from exploration to exploitation. The balance between these two modes is crucial: fitness-optimizing populations fail to explore enough to locate rare maximum-fitness individuals, while novelty-optimizing populations explore greatly but fail to exploit high-fitness individuals, and thus cannot derive any benefit from their exploration. Sel4Sel selection instead learns to explore in the beginning, then switch to strong exploitation once a sufficiently high-fitness individual is found.


\begin{wraptable}{r}{0.5\linewidth}
\center{
\begin{tabular}{|c||c|c|}\hline
Algorithm & Final Fitness & Final Novelty\\ \hline\hline
\textbf{Sel4Sel} & \textbf{32.0 $\pm$ 0.0} & 0.0 $\pm$ 0.0 \\
Underlying Fit. & 22.0 $\pm$ 4.82 & 0.0 $\pm$ 0.0 \\
Novelty & 6.54 $\pm$ 0.45 & 7.72 $\pm$ 0.55 \\
Min Crit. & 17.18 $\pm$ 1.07 & 0.43 $\pm$ 0.43 \\
Random Drift & 6.57 $\pm$ 0.57 & 3.5 $\pm$ 0.57 \\ \hline
\end{tabular}
}
\caption{\textbf{Deceptive Bits:} Average fitness and novelty of final populations, using various algorithms.} \label{table:deceptive_numbers}
\end{wraptable}

\textbf{Deceptive Bits.} In a final experiment, the Deceptive Bits domain presents a fitness landscape that is intentionally deceptive. As shown in Table \ref{table:deceptive_numbers}, populations optimizing for underlying fitness fall into one of many local maxima, resulting in low final fitness.

Figure \ref{fig:deceptive_fit} showcases how Sel4Sel learns to consistently achieves maximum fitness, even in the deceptive landscape. The crucial factor is how diversity affects a population’s ability to avoid local maxima. If a single individual optimizes greedily, it is likely to be trapped. However, if instead a population of diverse individuals is optimized, the odds are much higher that one of the individuals will go down the correct path towards maximal fitness. On the left side of Figure \ref{fig:deceptive_fit}, the novelty of the fitness-optimizing population is almost always at zero, indicating a homogenous population. Sel4Sel populations, on the other hand, display a high level of early novelty, corresponding to a more diverse population that is less prone to becoming trapped. In comparison to the Hashed Bits domain, there is not an explicit split between exploration and exploitation -- rather, Sel4Sel still selects for increasing fitness, but also maintains diversity by rewarding novel individuals. When a maximum-fitness individual is found, Sel4Sel again begins to fully exploit, driving novelty down to zero, but pushing average fitness to the maximum.

\subsection{What input metrics do strong Sel4Sel networks select for?}

\begin{table}[h]
\center{
\begin{tabular}{|c||c|c|c|c|c|}\hline
Sel4Sel Network & Underlying Fit. & Rank & Age & Novelty & Noise\\ \hline\hline
Convex Bits [Gen 1] & -0.113 & -0.024 & \textbf{0.213} & -0.821 & 0.075 \\
Convex Bits [Gen 100] & \textbf{0.954} & \textbf{0.946} & 0.279 & -0.968 & 0.018 \\
Convex Bits [Gen 500] & \textbf{0.970} & \textbf{0.986} & 0.292 & -0.986 & -0.051 \\
Convex Bits [Gen 1000] & \textbf{0.924} & \textbf{0.913} & 0.245 & -0.957 & -0.003 \\
Convex Bits [Gen 2000] & \textbf{0.950} & \textbf{0.965} & 0.289 & -0.958 & 0.148 \\ \hline

Hashed Bits [Gen 1] & 0.608 & 0.536 & -0.312 & \textbf{0.92} & 0.072 \\
Hashed Bits [Gen 100] & 0.179 & 0.238 & -0.023 & \textbf{0.725} & -0.046 \\
Hashed Bits [Gen 500] & \textbf{0.371} & 0.21 & 0.09 & 0.279 & -0.041 \\
Hashed Bits [Gen 1000] & \textbf{0.902} & \textbf{0.904} & 0.266 & -0.702 & -0.12 \\
Hashed Bits [Gen 2000] & \textbf{0.864} & \textbf{0.881} & 0.307 & -0.437 & -0.203 \\ \hline

Deceptive Bits [Gen 1] & -0.003 & 0.028 & -0.313 & \textbf{0.919} & -0.113 \\
Deceptive Bits [Gen 100] & 0.015 & 0.047 & 0.082 & \textbf{0.491} & -0.108 \\
Deceptive Bits [Gen 500] & 0.276 & 0.146 & -0.042 & \textbf{0.457} & -0.098 \\
Deceptive Bits [Gen 1000] & \textbf{0.966} & \textbf{0.958} & 0.291 & -0.92 & -0.115 \\
Deceptive Bits [Gen 2000] & \textbf{0.995} & \textbf{0.994} & 0.183 & -0.984 & 0.034 \\ \hline
\end{tabular}
}
\vskip 0.25cm
\caption{Pearson correlations between various metrics of an individual (Underlying Fitness, Rank, Age, Novelty, Noise) and their assigned internal fitness score. }\label{table:correlation}
\vspace{-0.5cm}
\end{table}

This work presents a key framework to gaining insight into strong evolutionary objectives: allow for the objectives themselves to evolve, then examine which objectives resulted in the strongest learning performance. In Table \ref{table:correlation}, the selective behavior of optimized Sel4Sel networks are examined, by presenting Pearson correlations between various input metrics and the resulting internal fitness scores.

Correlations on the Convex Bits domain show that underlying fitness and rank are the largest factors for internal fitness. Intuitively, this is comparable to a selection function that greedily favors high fitness individuals, a behavior that is shown to perform strongly in the domain. In domains with smooth fitness landscapes, greedy selection is the way to go. 

On both the Hashed Bits and Deceptive Bits domains, the largest factor in determining internal fitness early on is novelty. In both domains, some amount of exploration is required in order to locate the optimal solution. Notably, Sel4Sel networks have options for introducing variability into the population, e.g. selecting based on noise or for younger individuals, however results show that selecting for novelty is the strongest way to diversify. It is important to note that in both domains the pressure to diversify fades in later generations. This corresponds to the point where maximum fitness individuals have appeared in the population, and thus selection aims to remove lesser-fit individuals to maintain a high average. “First explore, then exploit” is displayed at work.

\section{Discussion and Conclusion}

Evolutionary algorithms are largely guided by their selective behavior, thus, is important for selection metrics to strike the right balance of exploration and exploitation. In this work, the novel algorithm Sel4Sel is introduced, which aims to \emph{automatically} discover neural-network-based selection metrics through a meta-evolutionary loop. On three bitstring domains with distinct fitness landscapes, Sel4Sel metrics match or outperform all baselines. Analysis of the learned metrics reveals that Sel4Sel consistently balances early exploration with efficient exploitation.

A key direction in this work is the perspective of using meta-learning to optimize evolutionary search. Given a distribution of tasks, there is always an optimal evolutionary objective which maximizes expected fitness of the final population, although it may be hard to locate. Sel4Sel provides a way to consider a wide distribution of possible objectives in order to discover an optimum. This framework is useful (1) as an empirical testing ground, since any consistently-appearing behavior is likely to generally useful, and (2) as a method to evaluate not only novel algorithms, but novel \emph{distributions} of algorithms, as Sel4Sel can discover the optimum within such distributions.

The limitations of Sel4Sel lie in the space over which meta-evolution takes place, namely all selection functions definable by a neural network. In some cases, the meta-evolutionary landscape may itself be deceptive, thus Sel4Sel would fail to converge to a strong selection function, even if such a function did exist. In other cases, adjusting the selection function of an evolutionary algorithm may not be enough for meaningful improvement, in which case similar frameworks must be adopted to meta-evolve other aspects of evolution, e.g. mutation, genetic operators, or population size.

Domains in this work were chosen to showcase solutions to distinct fitness landscapes, thus Sel4Sel networks were not aimed towards larger generalization. To encourage generalizability, more task variations can be introduced: the number of evolutionary generations could vary, or the fitness landscape could be randomly chosen each time. Even stronger would be to consider an open-ended task distribution where the fitness landscape is affected by peers, thus causing the fitness landscape to dynamically vary over generations. A fruitful direction may lie in investigating what evolutionary behaviors adaptively encourage learning in larger and larger task distributions.



\bibliographystyle{apalike}
\bibliography{refs}

\appendix

\end{document}